\pdfoutput=1
\documentclass[11pt,a4paper]{article}
\usepackage[hyperref]{acl2021}
\usepackage{times}
\usepackage{latexsym}

\usepackage{microtype}

\aclfinalcopy 

\usepackage{amsmath}
\newcommand{\norm}[1]{\left\lVert #1 \right\rVert}

\usepackage{tabularx}

\title{Marked Attribute Bias in Natural Language Inference}

\author{Hillary Dawkins \\
  University of Guelph, Ontario, Canada \\
  Vector Institute, Toronto, Ontario, Canada \\
  \texttt{hdawkins@uoguelph.ca} \\}

\date{}

\begin{document}
\maketitle
\begin{abstract}
Reporting and providing test sets for harmful bias in NLP applications is essential for building a robust understanding of the current problem.
We present a new observation of gender bias in a downstream NLP application: marked attribute bias in natural language inference. 
Bias in downstream applications can stem from training data, word embeddings, or be amplified by the model in use. However, focusing on biased word embeddings is potentially the most impactful first step due to their universal nature.  
Here we seek to understand how the intrinsic properties of word embeddings contribute to this observed marked attribute effect, and whether current post-processing methods address the bias successfully.
An investigation of the current debiasing landscape reveals two open problems: none of the current debiased embeddings mitigate the marked attribute error, and none of the intrinsic bias measures are predictive of the marked attribute effect.
By noticing that a new type of intrinsic bias measure correlates meaningfully with the marked attribute effect, we propose a new post-processing debiasing scheme for static word embeddings. The proposed method applied to existing embeddings achieves new best results on the marked attribute bias test set. See https://github.com/hillary-dawkins/MAB.
\end{abstract}

\section{Introduction}

Pre-trained distributed representations of words (a.k.a. word embeddings) are ubiquitous tools in natural language processing (NLP). 
Their utility is owing to the remarkable success in mapping semantic and syntactic relationships among words to linear relationships among real-valued vectors.  
For instance, analogy generation using vector addition on word embeddings (e.g. Tokyo is to Japan as Paris is to France) was taken to be an early measure of word embedding quality.  
In all kinds of related tasks, the vector space is known to encode semantic meaning surprisingly well \citep{pennington2014glove, Mik:2013:w2v, Mik:2013:linreg}. 
However, harmful gender-biased properties of word embeddings are also known to exist. 
Later is was observed that the same analogy generation property that produced the celebrated ``man is to king as woman is to queen" analogy  would also predict ``man is to programmer as woman is to homemaker" \citep{Tolga:2016}. This observation sparked interest in developing debiased word embeddings. 

Post-processing debiasing schemes are usually motivated by recognizing some intrinsic measure of bias in the embedding space, and then attempting to reduce that intrinsic bias. Early work (2016-2017) focused on the idea of a ``gender direction" vector within the embedding space, loosely defined as the difference vector between female and male attribute words.  
It was noted that any non-zero projection of a word onto the gender direction (termed direct bias) implied that the word was more related to one gender over another.
In the case of ideally gender-neutral words (e.g. doctor, nurse, programmer, homemaker), this was viewed as an undesirable property. 
The first debiasing methods, Hard Debias \citep{Tolga:2016} and Gender Neutral-GloVe \citep{Zhao2018:GNglove}, worked to minimize or eliminate the direct bias, and were shown to be successful in mitigating harmful analogies generated by word embeddings in relation to gender-stereotyped occupations.   

An influential critique paper by \citet{Gonen:2019} demonstrated that minimizing direct bias did not eliminate bias in the vector space entirely. Rather, words that tended to cluster together due to gender bias (e.g. nurse, teacher, secretary, etc.) would still cluster together in the nullspace of the gender direction. Furthermore, the original bias could be recovered by classification techniques using only the debiased word embeddings as input. These observations were termed cluster and recoverability bias.  

The next wave of debiasing methods (2019-present) focused on reducing cluster and recoverability bias while proposing new metrics to systematically quantify the indirect bias of the embedding space (e.g. the Gender-based Illicit Proximity Estimate, introduced by \citet{Kumar:2020:RAN}).  While these new debiasing schemes do reduce indirect bias in multiple ways, there is a general lack of
connection to downstream applications such as coreference resolution, natural language inference (NLI) and sentiment analysis. 

Current gender-bias evaluation tests (GBETs) in widespread use include the WinoBias test set \citep{Zhao:2018:WinoBias}, designed to measure bias in coreference resolution systems using stereotypical occupations as a probe, and the NLI test set \citep{Dev:2020:NLItest}, designed to measure stereotypical inferences again using occupations as the concept of interest.  
More commonly used evaluations include the Word Embedding Association Test (WEAT) \citep{Caliskan:2017:weat}, and the analogy generation test SemBias \citep{Zhao2018:GNglove}.  However these tests solely evaluate the vector properties of the word embeddings, without any connection to downstream applications. 
Adding to the library of downstream GBETs is essential in building a robust understanding of gender bias in NLP applications \citep{Sun:2019:review}.

Here we introduce a new observation of gender-biased predictions in a downstream task, namely “marked attribute bias” in natural language inference, and develop corresponding GBETs. Marked attribute bias refers to the language model’s tendency to predict that “person” implies “man” (the default attribute), while simultaneously understanding that “person” does not necessarily imply “woman” (the marked attribute). Marked attribute bias was found to exist on explicitly defined gender words (e.g. man, woman, etc.), and persist on implicit gender words (e.g. names) as well as latent gender-carriers (e.g. stereotypical occupations). 

An analysis of the currently available debiased embeddings reveals that none are able to successfully mitigate marked attribute bias. Furthermore, none of the currently proposed measures of intrinsic bias on the embedding space are predictive of the marked attribute effect.  We define a new measure of intrinsic bias that was found to correlate with the marked attribute effect better than any currently available metric. 
Using this insight, we introduce a new debiasing scheme: Multi-dimensional Information-weighted Soft Projection. Applying MISP to an existing debiased embedding achieves the lowest observed marked attribute bias error.  
\\

\textbf{Summary of main contributions:}

\begin{enumerate}
\item We present a new observation of gender bias in a downstream NLP application: marked attribute bias (MAB). The MAB test sets are made available in order to expand the current set of GBETs.
\item An analysis of current debiasing methods and current intrinsic bias measures finds that none sufficiently mitigate the error, and likewise none sufficiently explain the effect. This observation creates two new open problems.
\item We propose a new measure for quantifying intrinsic bias on the embedding space: Multi-dimensional Information-weighted Direct Bias (MIDB). This measure was found to correlate meaningfully with the marked attribute effect.
\item We introduce a new debiasing scheme: Multi-dimensional Information-weighted Soft Projection. MISP-debiased embeddings obtain new best performance on the MAB test set. 
\end{enumerate}

\section{Marked Attribute Bias in Natural Language Inference}

\subsection{Background: Natural Language Inference}

Natural language inference is one of the pillars of natural language understanding. 
It is the task of determining whether a hypothesis sentence is (neutral, entailed, or contradicted) with respect to a premise sentence. For example: 

\textbf{Premise:} A choir sings in the church. 

\textbf{Hypothesis:} The church is filled with the sound of singing. (Correct prediction: Entail) 

\citet{Dev:2020:NLItest} previously used NLI as a test case for gender bias with respect to occupations. For example, consider:   

\textbf{Premise:} A doctor prepared a meal. 

\textbf{Hypothesis 1:} A man prepared a meal. (N)

\textbf{Hypothesis 2:} A woman prepared a meal. (N)

This inference task essentially asks the question: is ``doctor" a subset of man/woman? I.e. if someone is a doctor, must they be a man? 
While both hypothesis sentences should receive a neutral prediction (as ``doctor" does not imply any specific gender), hypothesis 1 will more likely receive an entailment, while hypothesis 2 will more likely receive a contradiction, given biased word embeddings. The corresponding GBET was published by \citet{Dev:2020:NLItest} and contains 1936512 sentence pairs in the form [A \textbf{occupation} \textit{verb object}] $\rightarrow$ [A \textbf{gender word} \textit{verb object}].  Throughout this paper, we will use the notation [Sentence A] $\rightarrow$ [Sentence B] to mean that premise Sentence A is paired with hypothesis Sentence B.

\subsection{Observation of marked vs. default attribute bias}

Marked vs.~default attribute bias occurs whenever a default attribute (e.g. male, white, etc.) is assumed, and a marked attribute has to be explicitly stated or becomes a defining trait. In the context of the natural language inference task, consider the sentence pair:   

\textbf{Premise:} A person prepared a meal. 

\textbf{Hypothesis 1:} He prepared a meal. (N)

\textbf{Hypothesis 2:} She prepared a meal. (N) 

Due to the language model's\footnote{All NLI models mentioned throughout this paper are based on the Decomposable Attention Model \citep{parikh:2016:DAM} with intra-attention, trained on the Stanford Natural Language Inference training dataset \citep{bowman:2015:snli} (trained for 100 epochs; learning rate 0.025; weight decay 1e-5; dropout rate 0.2; 200 hidden units; approximately $10^4$ total model parameters).  All the code and data needed to reproduce results mentioned in this paper are available at https://github.com/hillary-dawkins/MAB.} tendency to predict that ``person" implies a male (default) attribute, the first hypothesis sentence will have a prediction probability vector shifted towards Entail. However the same language model would tend towards a Neutral prediction for the second hypothesis, recognizing that ``person" does not necessarily imply female (the marked attribute). 
To put it another way, this inference task essentially asks the question: is “person” a subset of man/woman? 
When presented with a masculine form, the model answers: yes (entailment), a person must be a man. When presented with a feminine form, the model answers: not necessarily (neutral), a female has an attribute (gender) that not all persons have.   
The name “Marked Attribute Bias” therefore derives from the observation that masculine forms are unmarked with respect to gender, whereas female forms carry a marked gender attribute. 

In this particular example, the model trained with (original) GloVe\footnote{Taken as the GloVe embeddings trained on the Common Crawl corpus for 840B tokens; available at https://nlp.stanford.edu/projects/glove/. Results were not found to vary significantly among undebiased embeddings.} word embeddings \citep{pennington2014glove} gives a probability distribution $(N, E, C)$ of (0.0538, \textbf{0.929}, 0.0177) for hypothesis 1 and (\textbf{0.687}, 0.238, 0.0750) for hypothesis 2.  

Note that although the MAB test construction appears similar to \citet{Dev:2020:NLItest}, it is actually measuring quite a distinct effect. The \citep{Dev:2020:NLItest} test set measures associations between gender and some concept of interest (occupations). The MAB test set measures something more general and pervasive; it measures how gender words carry meaning, independent of any concept of interest.

Achieving the correct prediction probability of $(N, E, C) = (1, 0, 0)$ on both sentences is difficult because it requires the language model to be attribute-aware (in this case gender-aware) while not using the gender attribute to alter predictions when it would be inappropriate to do so. 

\section{Analysis of the current situation}

In order to investigate the presence of systematic marked attribute bias in natural language inference, we construct three types of tests: bias on explicit gender words, implicit gender carriers, and latent gender carriers. We wish to understand the depth and persistence of the marked attribute effect, as well as how it is handled by current debiasing methods. Firstly we provide a brief description of the current debiasing methods to be analyzed. Next we provide details of the test sets and report results.

\subsection{Debiased embeddings}

Within the scope of this paper, we focus on post-processing techniques applied to static word embeddings. These types of methods are computationally inexpensive, easy to concatenate, and are independent of the base embedding. In addition, we include GN-GloVe, one of the highly cited retraining methods. Notationally, we specify embeddings as (base embedding).method. 
Where available, we use published debiased embeddings made available from the original authors of the corresponding method. Otherwise, we apply the method to the base GloVe embeddings. 
The methods we will analyze include:

\textbf{Hard Debias\footnote{https://github.com/tolga-b/debiaswe} (GloVe\footnote{\label{gloveZ} The base (undebiased) embeddings are GloVe trained on the 2017 January Wikipedia dump (vocab contains 322,636 tokens). Available at https://github.com/uclanlp/gn\texttt{\_}glove.}.HD)} \citep{Tolga:2016}: The subset of gender-neutral words are projected onto the nullspace of the gender direction $\vec{g}$. Gender-neutral words are made equidistant to pairs of words in a defined equalization set.  

\textbf{Gender-Neutral GloVe\footnote{https://github.com/uclanlp/gn\texttt{\_}glove} (GN-GloVe)} \citep{Zhao2018:GNglove}: Similar to hard debias, this method seeks to eliminate the direct bias. The embeddings are retrained from scratch using a modified version of GloVe’s original objective function. The gender information is sequestered to the final component of the word embedding. The gender-neutral portion of the word embedding is then defined as the first $d-1 = 299$ components, denoted \textbf{GN-GloVe$\pmb{(w_a)}$}. 

\textbf{Gender-Preserving\footnote{https://github.com/kanekomasahiro/gp\texttt{\_}debias} (GloVe\footnotemark[4].GP)} \citep{Kaneko:2019}: This method seeks to eliminate harmful gender bias while retaining as much useful semantic gender information as possible. 

\textbf{Double Hard Debias\footnote{https://github.com/uvavision/Double-Hard-Debias} (GloVe\footnotemark[4].DHD)} \citep{Wang:2020:DHD}: An extended version of the hard debias algorithm, based on the observation that frequency information encoded in the word embeddings convolutes the definition of the gender direction. Correctional pre-processing is applied prior to hard debiasing.

\textbf{Bias Alignment Model\footnote{https://github.com/anlausch/DEBIE} (GloVe\footnotemark[4].BAM)} \citep{Lauscher:2019}: Gender subspace matrices are defined by stacking explicit gender words. The projection that maps the embedding space to itself while approximately aligning the gender subspaces is learned and applied to all words. After alignment, gender information is not retained.

\textbf{Orthogonal Subspace Correction and Rectification\footnote{https://github.com/sunipa/OSCaR-Orthogonal-Subspace-Correction-and-Rectification} (GloVe\footnotemark[4].OSCaR)} \citep{Dev:2020:oscar}: 
The rationale is that linear projective methods are too aggressive in modifying the entire embedding space. 
OSCaR rectifies two concepts of interest (gender and occupations), such that these subspaces are orthogonal in the debiased space. 

\textbf{Iterative Nullspace Linear Projection\footnote{https://github.com/shauli-ravfogel/nullspace\texttt{\_}projection. The projection matrix computed for our base GloVe embeddings is available at https://github.com/hillary-dawkins/MAB.} (GloVe\footnotemark[4].INLP)} \citep{Ravfogel:2020:INLP}: Rather than defining a gender direction, INLP \textit{learns} the most informative decision boundary for classifying gendered and gender-neutral words. All words are projected to the nullspace of the gender subspace, and the process proceeds iteratively until gender information is sufficiently erased. 
A closely related method is the D$_4$ algorithm \citep{Davis:2020:D4}.

\textbf{Repulse Attract Neutralize Debias\footnote{https://github.com/TimeTraveller-San/RAN-Debias} (GloVe\footnotemark[4].RAN)} \citep{Kumar:2020:RAN}: Motivated by the persistence of implicit bias after debiasing through projective methods (observed as clustering and recoverability), RAN-debias attempts to address both direct bias and gender-based proximity bias.

\subsection{Explicit gender words test set and error definitions}

Firstly, we construct a test set where every sentence pair is of the form [A person \textit{verb object}] $\rightarrow$ [(A) \textbf{gender word} \textit{verb object}] (the correct inference is always neutral since a person can be of any gender).  Verbs ($n = 27$) and objects ($n = 184$) are paired to create $n = 1968$ unique premise sentences\footnote{Verbs and objects are taken from \citep{Dev:2020:NLItest} word lists (https://github.com/sunipa/On-Measuring-and-Mitigating-Biased-Inferences-of-Word-Embeddings) and are paired using the same pairing rules.}.  Gender words are taken to be \{man, woman, guy, girl, gentleman, lady, He, She\}, following \citep{Dev:2020:NLItest} with the addition of the pronouns,  for a total test set $S$ of $|S| = 15744$ sentence pairs where hypotheses represent binary genders evenly (denoted $S_M$, $S_F$,  $|S_M| = |S_F|$).   

For every hypothesis sentence in the test set, the ideal prediction probability vector is $(N, E, C) = (1, 0, 0)$. 
We could define the error on the test set as the average Euclidean distance from the ideal distribution:
\begin{equation}
\mathcal{E} = \frac{1}{|S|}\sum_{i \in S} \lVert (1,0,0) - (N, E, C)_i\rVert_2.
\end{equation}
This task, test set, and error definition are simple, and yet they encapsulate the central challenge of the debiasing field: to create attribute-aware (required to obtain the Neutral prediction) but attribute-unbiased embeddings. 

A weaker, but still potentially desirable, condition might be to minimize the effect of gender while not requiring that the model be gender-aware. Typically, this means that all hypotheses tend towards an Entail prediction, regardless of gender. We could define the error as the average distance between probability vectors between genders:
\begin{equation}
d = \frac{1}{2|S|} \norm{\sum_{i \in S_M}(N,E,C)_i - \sum_{j \in S_F}(N,E,C)_j  }_2.
\end{equation}
A gender-agnostic model could achieve zero error by this definition even with an accuracy of zero on the test set.  

Table 1 shows the results for this test set on all the embeddings of interest.  None of the debiased embeddings successfully mitigate the marked attribute error.
A similar test set shows that the effect persists on implicit gender words (e.g. names). Results are shown in the appendix.

\begin{table*}
\caption{Results of the marked attribute test set on \textbf{explicit gender words}. Due to varying results on gender nouns vs.~pronouns, results are shown separately for each case (M and F represent averages across the gender nouns).  Some debiased embeddings are able to eliminate the distance across pronouns (really by definition since $\vec{she} \approx \vec{he}$ in these cases), but none are able to eliminate differences between the gender nouns significantly. Even when differences between genders are minimized, distance from the ideal distribution (error $\mathcal{E}$) remains or increases. This highlights the challenge of creating gender-aware but not gender-biased embeddings.}
\begin{tabularx}{\textwidth}{l | Xrrr | Xrrr | rr}
\hline
Emb.method & Gender & N & E & C & Gender & N & E & C & $d$ & $\mathcal{E}$ \\  
\hline
GloVe & M &  0.7832 &0.1966 &0.0202& F   &0.9449 &0.0401 &0.0149 & 0.225 & 0.182 \\
          & he  &0.0982 &0.8838 &0.0180 &she &0.6549 &0.3137 &0.0315 & 0.797 & 0.865 \\
\hline
GloVe.HD & M &  0.8306& 0.1329 &0.0365& F  & 0.9269& 0.0499& 0.0232 & 0.128 & 0.155 \\
                & he  &0.2944 &0.6737 &0.0319 &she &0.5174 &0.4334 &0.0491 & 0.328 & 0.813 \\
\hline
GN-GloVe & M   &0.6339 &0.3402 &0.0259 &F   &0.9169 &0.0461 &0.0370 & 0.408 & 0.301 \\
               &he  &0.1767 &0.7968 &0.0265 &she &0.8223 &0.1405 &0.0373 & 0.921 & \textbf{0.688} \\
\hline
GN-GloVe($w_a$) & M &  0.8446& 0.1254& 0.0300& F &  0.9211& 0.0395 &0.0394 & \textbf{0.115} & \textbf{0.149} \\
                     & he & 0.1430 &0.8266 &0.0304 &she& 0.4237 &0.5367 &0.0396 & 0.404 & 0.990 \\
\hline 
GloVe.DHD & M   &0.7013 &0.2685 &0.0302 &F   &0.9282 &0.0510 &0.0209 & 0.315 & 0.247 \\
                   & he  &0.1566 &0.8187 &0.0247 &she &0.1597 &0.8139& 0.0264 & 0.006 & 1.173 \\
\hline
GloVe.GP  &M   &0.6172 &0.3521 &0.0306 &F   &0.8777 &0.0693 &0.0530 & 0.385 & 0.336 \\
                 & he  &0.2443 &0.7262 &0.0295 &she &0.6481& 0.3040 &0.0480 & 0.585 & 0.758 \\
\hline
GloVe.BAM  & M   &0.7983 &0.1703 &0.0314 &F   &0.9329 &0.0447 &0.0224 & 0.184 & 0.175 \\
                    & he  &0.1625 &0.8083 &0.0292 &she &0.6752 &0.2878 &0.0369 & 0.731 & 0.800 \\
\hline
GloVe.OSCaR  & M &  0.8233& 0.1572& 0.0195& F   &0.9431 &0.0400 &0.0169 & 0.168 & 0.154 \\
                        & he & 0.1482 &0.8292 &0.0226 &she &0.8428 &0.1278 &0.0294 & 0.987 & 0.697 \\
\hline
GloVe.RAN    & M   & 0.8055 &0.1686 &0.0260 &F   &0.8994 &0.0701 &0.0305 & 0.136 & 0.193 \\
                     & he  &0.1939 &0.7811 &0.0250 &she &0.5962 &0.3420 &0.0618 & 0.597 & 0.828 \\
\hline
GloVe.INLP & M   &0.8298 &0.1537 &0.0166 &F   &0.9204 &0.0633 &0.0164 & 0.128 & 0.167 \\
                   & he  &0.1081 &0.8710 &0.0209 &she &0.1119 &0.8672 &0.0209 & \textbf{0.005} & 1.244 \\
\hline 
\end{tabularx}
\label{tab:AllEmbs}
\end{table*} 

\subsection{Latent gender carriers: Stereotyped occupations}

\begin{table*}
\caption{Results of marked attribute test set on \textbf{stereotypical occupations}.  Each (N,E,C) probability vector is averaged over the 1968 unique premise sentences and the gender attribute words from each category (M or F) ($n = 31,488$ sentences for each gender). Smaller distances between the M and F vectors indicate less gender bias. The significance of the difference was evaluated using a permutation test; the alternate distance $d^*$ is computed for 10,000 randomly sampled partitions of the occupations into two groups. The significance value is the proportion of these samples to generate a distance $d^* > d$. This gives us an idea of whether the defined partition, based on gender, is a meaningful grouping. Smaller significance values indicate that the defined partition is non-random with respect to the distance.}
\begin{tabularx}{\textwidth}{l | X | X | rr}
\hline 
Emb.method & M attribute (N, E, C) &  F attribute (N, E, C)  & Distance $d$ & Significance \\
\hline 
GloVe& (0.6000, 0.3350, 0.0650) &(0.7378, 0.1711, 0.0910)& 0.216  & 0.0001 \\
GloVe.HD & (0.4975, 0.4500, 0.0525) &(0.6075, 0.3357, 0.0568) &0.159 & 0.0408 \\
GN-GloVe & (0.5026, 0.4434, 0.0540) &(0.7126, 0.2036, 0.0838) &0.320 &  0.0000 \\
GN-GloVe$(w_a)$ & (0.5309, 0.3915, 0.0776)& (0.6197, 0.2771, 0.1032)& 0.147 &  0.0478 \\
GloVe.DHD & (0.5285, 0.4126, 0.0589) &(0.6513, 0.2811, 0.0676) &0.180 & 0.0038 \\
GloVe.GP & (0.5016, 0.4380, 0.0604) &(0.6479, 0.2639, 0.0882) &0.229 & 0.0010 \\
GloVe.BAM & (0.6293, 0.3077, 0.0630) &(0.7116, 0.1972, 0.0912) &0.141 & 0.0060 \\
GloVe.OSCaR & (0.5577, 0.3901, 0.0522)& (0.6789, 0.2400, 0.0812)& 0.195 &  0.0036 \\
GloVe.RAN & (0.5393, 0.3933, 0.0674) &(0.5924, 0.3026, 0.1050) &0.112 &  0.0477 \\
GloVe.INLP & (0.5065, 0.4197, 0.0739) &(0.5465, 0.3949, 0.0587) &0.050 & 0.6595 \\
\hline 
\end{tabularx}
\label{tab:Occs}
\end{table*}

Next, we would like to check if the gender-induced marked attribute bias can affect entities which should be gender neutral, but turn out to be hidden carriers of a gender attribute (e.g. stereotypical occupations). The same template [A person \textit{verb object}] $\rightarrow$ [A/An \textbf{occupation}  \textit{verb object}] was used with the common vocabulary. Stereotypical occupations ($n=32$) were sourced from \citet{Tolga:2016}, and the SemBias test set. Examples are (doctor, engineer, boss, etc.~vs.~nurse, maid, homemaker, etc.). In total there are 62,976 sentence pairs in the test set\footnote{\label{wordset}The exact word set used to produce these results is available at https://github.com/hillary-dawkins/MAB.}. 

Results are shown in Table 2.  A permutation test is used to check if dividing the occupations into groups according to gender stereotypes produces a significant difference in the probability vectors (rather than dividing them randomly). As shown, the marked attribute effect persists on stereotypical occupations, especially on original embeddings. This is an important result because it highlights that unintended behaviour can appear in unexpected places due to a latent attribute. 
Previously, GBETs have focused on how explicit gender words are treated under biased models. To our knowledge, this is the first GBET designed to analyze unintended behaviour on a latent attribute carrier.

Note that this task is easier to correct than the explicit gender words because occupation words have defining characteristics beyond gender. That is, a debiasing method such as Iterative Nullspace Projection can perform well by removing gender information entirely.
This does not mean that the challenge of having a gender-aware but gender-unbiased embedding is solved, but it does provide evidence that latent gender effects can be mitigated using linear projective methods. 
The full extent of latent biased-attribute effects and possible mitigation strategies should be investigated further.

\section{Intrinsic bias measures}

How to define bias on an embedding space remains an active area of study.
In general, we seek to understand how the intrinsic or geometric properties of an embedding space translate to real observable bias in downstream tasks. Intrinsic properties are easy to compute quickly, whereas computing performance on downstream tasks requires us to train new models for every case. Understanding of the correlations between the two gives insight on how word embeddings should be debiased. 

As a case study, let us focus on the marked attribute error $\mathcal{E}$ on the explicit gender words (shown in Table 1). Recall that this measure of bias is of interest because zero error corresponds to the gold standard: having an attribute-aware model, while simultaneously not using the gender attribute to make inappropriate inferences. 
In this section, we look at 5 existing intrinsic bias measures: Direct Bias, Clustering, Recoverability, Gender-based Illicit Proximity Estimate (GIPE), and SemBias. We will investigate whether any of these measures are predictive of the marked attribute effect.

Recall that direct bias was the first measure to be proposed; it simply measures the average projection of word vectors onto a predefined gender direction. 
Early methods (i.e. Hard Debias and GN-GloVe) defined bias in the embedding space completely as direct bias.
The idea of clustering and recoverability refer to a classifier's ability to correctly reassign gender labels to words, even after debiasing methods have been applied. 
\citet{Gonen:2019}'s observation of clustering and recoverability sparked new interest in defining metrics for indirect bias on the embedding space. 
Although clustering and recoverability do not provide well-defined measures of bias given an embedding space (as they depend training implementation - though they could be said to provide a lower bound), many new debiasing proposals will cite reduced clustering as a positive result. The effect on downstream applications is not well understood as of yet. 
The Gender-based Illicit Proximity Estimate (GIPE) measures the extent of undue proximities in the embedding space due to a pervasive gender attribute.
Lastly, the SemBias analogy test set measures whether gender-biased analogies exist within the embedding space based on vector arithmetic properties. 

Implementation details for each measure as well as the experimental set of embeddings ($n = 16$) are given in the appendix. 
The average Direct Bias on the embedding space was found to have a Pearson correlation coefficient of 0.104 with the marked attribute error.  The Clustering $v$-measure\footnote{With cluster size $n = 1500$ (which lead to the highest observed correlation); see appendix.} \citep{rosenberg:vmeas} achieved a correlation coefficient of 0.184.  Recoverability was attempted using an SVM with a linear decision boundary, an SVM with a non-linear (radial basis function) kernel, logistic regression, and a simple 1-hidden-layer fully-connected network.  All recoverability correlation results were comparable, but the best coefficient of 0.223 was achieved by logistic regression.  The GIPE\footnote{Using an indirect bias threshold of $\theta = 0.05$, and number of nearest neighbours $n = 100$.} had a correlation coefficient of 0.432. The SemBias\footnote{The SemBias score was taken as the proportion of analogy examples in the test set for which the embedding space returns the correct definitional analogy.} test set had a correlation coefficient of 0.091.
The full correlation matrix between all intrinsic bias measures can be found in the appendix. 
The results suggest that the marked attribute effect is not well correlated with any present notion of intrinsic bias, therefore we do not have a good understanding of how the word embedding properties contribute to this type of observable bias.

In seeking a potential solution, we make note of a new intrinsic bias measure, multi-dimensional information-weighted direct bias (MIDB), found to have a more meaningful correlation of 0.667 with the marked attribute error. We define the MIDB of a particular word $\vec{x}$ to be a weighted average over inner products with basis vectors of a multi-dimensional gender subspace:
\begin{align}
\text{MIDB}_d(x) = \sum_{i=1}^d a_i \langle g_i |  x \rangle
\end{align}
where $\{ g_i\}$ form an orthonormal basis for the gender subspace, here defined as the first $d$ principal components summarizing difference vectors $\{ \delta_{jk}\}$. The difference vectors are taken as all pairwise differences\footnote{Using all pairwise differences creates a matrix with rank much less than the dimension of the matrix, however the rank is still much larger than $d$ (the number of principal components to extract) so it doesn't cause a problem.} between vectors in defined gender sets (here common names were used\footnotemark[13]): $\{ \delta_{jk}\} = \vec{f}_j - \vec{m}_k$, $f_j \in F_\text{names}$, $m_k \in M_\text{names}$ ($|M_\text{names}| = |F_\text{names}| = 100$).
The weighting $a_i$ is the proportion of variance explained by the $i^{th}$ principal component,
and $d$ is a hyperparameter controlling the number of dimensions to keep\footnote{On our set of experimental embeddings, $d = 4$ was empirically found to produce the 0.7 correlation result.}.

New proposals for defining a gender direction or subspace potentially have far reaching consequences in the landscape of intrinsic bias measures and their related debiasing schemes. In fact all of Clustering, Recoverability, GIPE, and SemBias use the classic uni-dimensional gender direction $\vec{g}$ within their definitions. 
The weak observed correlation between DB and MIDB suggests that these subspaces are independent. 
Swapping in a uniquely informative gender subspace to the existing indirect measures would produce a new family of intrinsic bias measures.
The observed utility of names in defining a meaningful gender subspace is encouraging because it opens an obvious avenue for this method to be applied to attributes of interest beyond gender (e.g. race or ethnicity). 

\section{Multi-dimensional information-weighted soft projection}

\begin{table*}
\caption{Results for word similarity and analogy benchmarks. Results on the word analogy tasks are reported as percentage accuracy. Results on the word similarity tasks are reported as a Spearman correlation ($\times$ 100). Application of MISP does not alter the overall quality of word embeddings as measured by these classic test sets.}
\begin{tabularx}{\textwidth}{l | rrrr| rrrr}
\hline 
Embedding.method & Sem &  Syn  & Google-Total & MSR & RG & MTurk & MEN & SL999 \\
\hline 
GloVe &  80.48 & 62.76 & 70.80 & 51.49 & 75.29 & 64.27 & 72.19 & 34.86   \\
GloVe.MISP &  80.49 & 62.81 & 70.84 & 51.51 & 76.06 & 64.32 & 72.41 & 35.04   \\
GN-GloVe & 77.62 & 61.60 & 68.87 & 49.29 & 74.11 & 66.36 & 74.49 & 37.12  \\
GN-GloVe$(w_a)$ & 77.68 & 61.56 & 68.87 & 49.38 & 75.46 & 66.55 & 74.72 & 37.53  \\
GN-GloVe$(w_a)$.MISP & 77.68 & 61.59 & 68.89 & 49.26 & 75.49 & 66.45 & 74.76 & 37.60  \\
\hline 
\end{tabularx}
\label{tab:Vanilla}
\end{table*}

In this section we motivate the above search for an informative intrinsic bias measure.  As discussed, a greater understanding of how embedding properties influence observed bias can inform new debiasing techniques. Translating the idea of MIDB into a debiasing scheme yields Multi-dimensional Information-weighted Soft Projection (MISP). 

In this debiasing procedure, we project all words into the nullspace of the multi-dimensional gender subspace, proportional to our belief that certain dimensions actually encode the latent idea of gender:
\begin{align}
\vec{w}_{deb} = \vec{w} - \sum_{i = 1}^{d} a_i \langle{g_i}|w\rangle |g_i\rangle
\end{align}
where $\vec{w}$ is the input embedding, $\vec{w}_{deb}$ is the debiased output embedding, and all other quantities are defined as in eq. (3).

As shown in Table 1, the GN-GloVe$(w_a)$ embeddings are currently the top performers on the explicit gender words test set, as measured by either error $\mathcal{E} = 0.149$, or distance $d = 0.115$.
Applying MISP to GN-GloVe$(w_a)$ embeddings (denoted GN-GloVe$(w_a)$.MISP), we achieve an error on the explicit gender words test set of $\mathcal{E} = 0.1107$, a 26\% error reduction over the previous best. 
The distance $d$ between genders is reduced to $d = 0.08744$, a 21\% reduction over the previous best.
Successful concatenation suggests that this technique is distinct, and independently useful, from techniques that seek to minimize the traditional direct bias (including GN-GloVe).  This observation is consistent with the weak observed correlation between direct bias and MIDB$_4$ on the experimental set of embeddings.

Computing the intrinsic bias measures Clustering, Recoverability, GIPE and SemBias on the newly created embedding space GN-GloVe$(w_a)$.MISP (compared to the base GN-GloVe$(w_a)$), we observe a clustering $v$-score of 0.498 (previously 0.497)\footnote{Where clustering size $n = 1500$.}, a recoverability accuracy of 0.992 (previously 0.993)\footnote{This is the highest accuracy achieved by any of the four classification methods tested; implementation details are in the appendix.}, a GIPE of 0.1169 (previously 0.1173)\footnote{Computed with indirect bias threshold $\theta = 0.03$, and number of nearest neighbours $n = 100$.}, and a SemBias score of 0.938 (previously 0.938)\footnote{Reported as the proportion of samples in the full test set to return the definitional analogy; higher scores are better.}.  
The MISP method did not reduce bias by any of these measures, although this is not particularly surprising as it was designed to address the marked attribute effect (through MIDB).  It is encouraging however that none of these bias measures were increased. In other words, there is no expected trade-off between the reduced marked attribute error and any previous debiasing work that relied on these measures. The SemBias result informs us that MISP did not reintroduce any harmful biased analogies, for example. 

For reference, if we apply the analogous multi-dimensional hard debias method (i.e. equation (4) where all weights $a_i$ are set to 1), the output embeddings GN-Glove$(w_a)$.MHD do not successfully mitigate the marked attribute effect ($\mathcal{E} = 0.1501$, $d = 0.1603$). This suggests that the soft nature of the projection is a key ingredient.  

Furthermore, we provide some evidence that specifically the information weighting of the soft projection is a good ingredient as follows.  Recall that we are attenuating components of each basis vector according to our belief in that vector as a good gender direction. 
The basis vectors are defined to be the first $d$ principal components, weighted by their corresponding variance explained. Therefore the first basis vector receives the greatest weight and so on. 
To test the significance of this decision, we define alternative debiased embeddings by applying MISP where the weights get reassigned to the ``wrong" vector (for $d=4$, we have 23 alternative pairings). 
We observe that \textbf{none} of the 23 alternatives obtain an error $\mathcal{E}$ less than the ``true" implementation of MISP.  This suggests that weighting the components by order of information is a good ingredient. Values of $\mathcal{E}$ for the alternate embeddings can be found in the appendix. Model parameters for each case are made available in order to reproduce this argument on any extended version of the MAB test set. 

Information weighting is an interesting idea because it could be applied to either defined or learned gender subspaces alike. For instance, if the basis vectors of a gender subspace are taken as the iteratively learned linear decision boundaries (as in INLP), we could investigate weighting each dimension by the accuracy $acc_i$ of classification on each iteration, as $a_i = (1-2acc_i)$. In this way, dimensions receive weights proportional to their ability to predict gender information. 
When accuracy reaches 0.5, no gender information remains, the learned decision boundary is meaningless, and the basis vector receives zero weighting.  

Finally, as with any debiasing method, we wish to verify that application of the method has not damaged the overall embedding quality.  We assess the MISP embeddings on a handful of classic analogy and word semantic similarity benchmarks. The word similarity benchmarks measure how closely the word embeddings capture similarity between words compared to human annotation. We use the following datasets: RG \citep{RG}, MTurk \citep{MTurk}, MEN \citep{MEN}, and SimLex999 \citep{simlex}. The word analogy task measures how well the word embeddings capture semantic and syntactic relationships among words as vector properties. We report on the Google \citep{Mik:2013:goog}, and MSR \citep{Mik:2013:linreg} test sets.  Results were obtained following the word embedding benchmark package\footnote{https://github.com/kudkudak/word-embeddings-benchmarks} \citep{Jastrzebski:2017}.  As shown, application of MISP does not alter the overall word embedding quality.

\section{Conclusion}

This paper highlights a new observation of gender bias in a downstream setting: marked attribute bias in natural language inference. 
The current inference is that ``person" implies male, while ``person" does not imply female. 
Consequently, this inference is being baked into our models of natural language understanding. 
The effect was shown to persist on explicitly defined gender words and on latent gender-attribute carriers.
Based on an assessment of the current debiasing landscape, none of the current debiasing methods satisfactorily mitigate the marked attribute error, and furthermore none of the intrinsic bias measures are useful at predicting the marked attribute effect. 

By noticing a more meaningful correlation with a newly identified intrinsic bias measure, we propose a new debiasing scheme: multi-dimensional information-weighted soft projection (MISP).
This method introduces several concepts, including the use of a multi-dimensional defined gender subspace. 
Previously, the concept of a defined gender subspace always appeared as a single dimension. The iterative nullspace projection method implicitly uses higher learned dimensions, however this requires learning a new decision boundary at every iteration, subject to the implementation of a training procedure. Furthermore, the learned dimensions were not used to define any bias metric, they were strictly used operationally for the debiasing procedure. MISP also introduces the idea of a soft or partial projection, where weights are informed by some measure of the dimension’s ability to capture the intended latent concept of a gender direction. Both of these ideas could be further explored and extended to create new notions of indirect bias, which in turn could inform more sophisticated debiasing procedures. 

Multi-dimensional information-weighted soft projection applied to GN-GloVe$(w_a)$ produces new debiased embeddings that achieve the lowest error on the marked attribute bias test set, a 26\% reduction over the previous best, and a 45\% reduction over the original undebiased embeddings. Error reduction on this test set is thought to encapsulate the overall goal of producing gender-aware but gender-unbiased embeddings. Therefore, this method and its composite ingredients warrant further investigation. 
Each of the marked attribute bias test sets are made available for further exploration and iteration on these ideas. 

\section*{Acknowledgements}
We thank Daniel Gillis, Judi McCuaig, Stefan Kremer, and Graham Taylor for their insightful comments and discussions. We thank the anonymous reviewers for their comments and suggestions which improved the final version of this manuscript. This work is financially supported by the Government of Canada through the NSERC CGS-D program (CGSD3-518897-2018). 

\bibliographystyle{acl_natbib}
\bibliography{acl2021}

\appendix

\section{Implicit gender carriers: Names}

\begin{table*}
\caption{Results of marked attribute test set on \textbf{names}.  Each (N,E,C) probability vector is averaged over the 1968 unique premise sentences and the gender attribute words from each category (M or F) ($n = 62,976$ sentences for each gender). Smaller distances between the M and F vectors indicate less gender bias. The significance of the difference was evaluated using a permutation test; the alternate distance $d^*$ is computed for 10,000 randomly sampled partitions of the names into two groups. The significance value is the proportion of these samples to generate a distance $d^* > d$. This gives us an idea of whether the defined partition, based on gender, is a meaningful grouping. Smaller significance values indicate that the defined partition is non-random with respect to the distance.}
\begin{tabularx}{\textwidth}{l | X | X | rr}
\hline 
Emb.method & M attribute (N, E, C) &  F attribute (N, E, C)  & Distance $d$ & Significance \\
\hline 
GloVe & (0.4657, 0.4766, 0.0577) & (0.7283, 0.1598, 0.1120)& 0.415 & 0.0000 \\ 
GloVe.HD &(0.5745, 0.3547, 0.0708)& (0.6685, 0.2760, 0.0555)& 0.124 & 0.0140 \\
GN-GloVe &(0.4619, 0.4713, 0.0668) &(0.7209, 0.1906, 0.0885) &0.383 &  0.0000 \\
GN-GloVe$(w_a)$& (0.5882, 0.2878, 0.1240)& (0.6662, 0.2321, 0.1017)& 0.098 &  0.0241 \\
GloVe.DHD &(0.4731, 0.4464, 0.0805) &(0.5690, 0.3529, 0.0780) &0.134 &  0.0192 \\
GloVe.GP &(0.5488, 0.3761, 0.0751) &(0.7470, 0.1677, 0.0853) &0.288 & 0.0000 \\
GloVe.BAM &(0.5941, 0.3424, 0.0635)& (0.7698, 0.1628, 0.0674)& 0.251 &  0.0000 \\
GloVe.OSCaR &(0.6012, 0.3149, 0.0839)& (0.7191, 0.2020, 0.0789) &0.163 &  0.0001 \\
GloVe.RAN &(0.5295, 0.3865, 0.0839) &(0.6920, 0.2151, 0.0929) &0.236 & 0.0000 \\
GloVe.INLP &(0.5091, 0.4042, 0.0867)& (0.5447, 0.3639, 0.0914) &0.054 & 0.4049 \\
\hline 
\end{tabularx}
\label{tab:Names}
\end{table*}

As mentioned in the main text, we check if the marked attribute effect will persist through implicit gender words. These are words with no gender attribute by definition, but are usually associated with a specific binary gender (e.g. names). This test set uses a similar template: [A person \textit{verb object}] $\rightarrow$ [\textbf{Name} \textit{verb object}], using the same (verb, object) vocabulary as above. Names\footnote{\label{wordset}The exact word set used to produce these results is available at https://github.com/hillary-dawkins/MAB.} ($n = 64$) are sourced from the most common names of the previous decade in the US, according to the Social Security Administration\footnote{https://www.ssa.gov/oact/babynames/}. In total there are $n = 125,952$ sentence pairs in the test set. 

Results are shown in Table 4. In short, the same effect is observed on names, especially on the original embeddings. 
A permutation test was used to check whether the stratification of names by gender was a non-random division according to the observed bias. 

\section{Intrinsic bias measures and correlations}
Please refer to tables 5 and 6.

\begin{table*}
\caption{Pearson correlation matrix between intrinsic bias measures (and marked attribute error) on the experimental set of embeddings. MIDB obtains the highest correlation with the marked attribute error $\mathcal{E}$; the GIPE was also observed to have a weak correlation. Recoverability bias is most related to the direct bias. The sub-matrix among the SemBias results indicate that trade-off is mostly happening between ``definitional" and ``other" analogies. }
\begin{tabularx}{\textwidth}{l | llllllll}
\hline 
 & DB$_{vt}$ & MIDB & Clus:$v_{1500}$ & Rec:LR & SB{def} & SB{stereo} & GIPE:0.03 & $\mathcal{E}$\\
\hline 
DB$_{vt}$	&1&	-0.166 &	0.694 &	0.814 &	0.161 &	-0.045 &	0.350 &	0.104 \\
MIDB	& &	1&	-0.145 &	-0.020 &	-0.273 &	0.005 &	-0.003 &	0.667 \\
Clus:$v_{1500}$ &	&	&1	&0.776	&0.600	&-0.185 	&0.271 &	0.184 \\
Rec:LR	&	& &	&	1&	0.786 &	-0.390 &	0.290 &	0.223 \\
SB{def}	&	& &	&	&	1&	-0.693 &	0.304 &	0.091 \\
SB{stereo}	& &	&	&	&	&	1&	-0.487 &	-0.270 \\
GIPE:0.03&	&	&	& 	&	& &	1	& 0.432 \\
$\mathcal{E}$&	&	&	& &	&		& &	1\\
\hline 
\end{tabularx}
\label{tab:Names}
\end{table*}

\begin{table*}
\caption{Intrinsic bias measures of interest on the experimental set of embeddings.  There are two base (undebiased) embeddings, word2vec and GloVe. All other embedding spaces are obtained by applying a debiasing method, where each method found here is described in the main text. Implementation notes: \\
 \textbf{DB and MIDB:} The direct bias (DB) and the new multi-dimensional information-weighted direct bias (MIDB) are average measures over a specific (ideally gender-neutral) vocabulary $V_t$.  $V_t$ ($n=46960$) is defined by taking the 50,000 most frequent words in the common vocabulary between word2vec and GloVe, filtering out punctuation, numbers, and removing the gender-specific word set $V_s$ ($n = 1622$), defined as the union of gender-specific word sets used in previous works \cite{Tolga:2016, Zhao2018:GNglove}.  DB is defined as the projection onto a gender direction, here taken to be the $\vec{she} - \vec{he}$ direction. For debiasing methods that promote $\vec{she} \approx \vec{he}$, the DB is not well defined (although it can be computed numerically, it is unstable). We leave these cases as NA rather than a spurious numerical value.\\
\textbf{Clustering:} The clustering experiment follows \cite{Gonen:2019} in taking the $n \in [500,1500]$ ``most biased" words in the original embedding space (according to their projection on the $\vec{she} - \vec{he}$ axis), and then applying $k$-means ($k=2$) clustering on the words in the debiased embedding space. Bias is reported as the either the clustering accuracy or the $v$-measure (only $n = 1500$ shown here with $v$-measure).  \\
\textbf{Recoverability:} Similarly, the dataset ($n=5000$) is taken to be the most biased words in the original embedding space, where bias labels are assigned according to the projection on the gender direction ($n=2500$ taken from each class).  Several classifiers (SVM with a linear decision boundary, SVM with an RBF kernel, logisitic regression, and a simple fully-connected 1-hidden layer network) were trained on $20\%$ of the dataset with balanced classes. Recoverability bias is reported as the accuracy of classification on the remaining test set (only logisitic regression shown here).       \\
\textbf{SemBias:} The SemBias analogy test set is available from \cite{Zhao2018:GNglove}. The set contains $n = 440$ tuples of possible analogies $(\vec{a}, \vec{b})$: 1 definitional analogy (e.g. king, queen), 1 stereotypical analogy (e.g. doctor, nurse), and 2 other analogies (e.g. cup, plate). For every sample, the best analogy is selected as the one to maximize $cos(\vec{he}-\vec{she}, \vec{a}-\vec{b})$. Bias is reported as the proportion of samples to return a definitional analogy, a stereotypical analogy, and an ``other" analogy. (Only definitional and stereotypical shown here.) \\
\textbf{GIPE:} The gender-based illicit proximity bias (GIPE) (see \cite{Kumar:2020:RAN} for details) was computed with $n = 100$ nearest neighbours for each word, with an indirect bias threshold of $\theta \in [0.03, 0.05]$ following \cite{Kumar:2020:RAN}. (Only $\theta = 0.03$ shown here.) \\
Full results, plus all code, embedding files, and word sets needed to replicate these results are available at https://github.com/hillary-dawkins/MAB.}
\begin{tabularx}{\textwidth}{l | llllllll}
\hline 
Emb.method & DB$_{vt}$ & MIDB & Clus:$v_{1500}$ & Rec:LR & SB{def} & SB{stereo} & GIPE:0.03 & $\mathcal{E}$\\
\hline 
w2v & 0.052 & 0.023 & 0.933 & 0.992 & 0.830 & 0.134 & 0.021 & 0.206 \\ 
w2v.HD & 0.000 & 0.007 & 0.440 & 0.887 & 0.759 & 0.114 & 0.014 & 0.163 \\ 
w2v.DHD & NA & 0.025 & 0.271 & 0.881 & 0.295 & 0.373 & 0.014 & 0.164 \\ 
w2v.BAM & 0.061 & 0.038 & 0.844 & 0.974 & 0.814 & 0.136 & 0.023 & 0.131 \\ 
w2v.OSCaR & 0.050 & 0.024 & 0.928 & 0.993 & 0.830 & 0.134 & 0.021 & 0.188 \\ 
GloVe & 0.055 & -0.032 & 0.984 & 1.000 & 0.802 & 0.109 & 0.115 & 0.198 \\ 
GloVe.HD & 0.000 & -0.004 & 0.302 & 0.927 & 0.786 & 0.130 & 0.070  & 0.155 \\
GN-GloVe & 0.038 & 0.172 & 0.588 & 0.999 & 0.977 & 0.014 & 0.141 & 0.301 \\
GN-GloVe$(w_a)$& 0.068 & -0.096 & 0.497 & 0.989 & 0.939 & 0.011 & 0.117 & 0.149 \\
GloVe.DHD & NA & 0.201 & 0.258 & 0.903 & 0.250 & 0.123 & 0.064 & 0.247 \\
GloVe.GP & 0.059 & 0.068 & 0.996 & 1.000 & 0.843 & 0.080 & 0.145 & 0.336 \\
GN-GloVe.GP & 0.036 & 0.006 & 0.601 & 0.999 & 0.984 & 0.011 & 0.118 & 0.179 \\
GloVe.BAM & 0.068 & -0.019 & 0.964 & 0.999 & 0.775 & 0.145 & 0.137 & 0.175 \\
GloVe.OSCaR & 0.056 & -0.012 & 0.984 & 1.000 & 0.814 & 0.102 & 0.117 & 0.154 \\
GloVe.RAN & 0.044 & -0.001 & 0.419 & 0.951 & 0.927 & 0.011 & 0.040 & 0.193 \\
GloVe.INLP & NA & -0.001 & 0.015 & 0.660 & 0.198 & 0.160 & 0.080 & 0.167 \\
\hline 
\end{tabularx}
\label{tab:Names}
\end{table*}

\section{Alternate weighted embeddings}

As discussed in the main text, we compute the error $\mathcal{E}$ on the explicit gender words test set for alternate soft-weighted embeddings. The alternate embeddings are created by permuting the weights to be matched with the incorrect basis vectors. For example, the permutation denoted 1243 means that weight $a_1$ is applied to basis vector $\vec{g_1}$, $a_2$ to $\vec{g_2}$, $a_4$ to $\vec{g_3}$, and $a_3$ to $\vec{g_4}$. Results for all alternate permutations and their errors are as follows: (permutation = 1243, $\mathcal{E}$ = 0.1574), (1324, 0.2331), (1342, 0.1919), (1423, 0.1330) (1432, 0.1487) (2134, 0.1273) (2143, 0.1565) (2314, 0.2289) (2341, 0.1963) (2413, 0.1639) (2431, 0.1287) (3124, 0.1951) (3142, 0.1694) (3214, 0.2813) (3241, 0.1602) (3412, 0.2110) (3421, 0.1732) (4123, 0.1945) (4132, 0.1764) (4213, 0.1879) (4231, 0.1363) (4312, 0.1241) (4321, 0.1435).   

\end{document}